# Non-Linearities Improve OrigiNet based on Active Imaging for Micro Expression Recognition

Monu Verma
Dept. of Computer Science and Engineering
Malaviya National Institute of Technology
Jaipur, India
monuverma.cv@gmail.com

Santosh Kumar Vipparthi
Dept. of Computer Science and Engineering
Malaviya National Institute of Technology
Jaipur, India
skvipparthi@mnit.ac.in

Girdhari Singh
Dept. of Computer Science and Engineering
Malaviya National Institute of Technology
Jaipur, India
gsingh.cse@mnit.ac.in

*Abstract*— Micro expression recognition (MER)is a very challenging task as the expression lives very short in nature and demands feature modeling with the involvement of both spatial and temporal dynamics. Existing MER systems exploit CNN networks to spot the significant features of minor muscle movements and subtle changes. However, existing networks fail to establish a relationship between spatial features of facial appearance and temporal variations of facial dynamics. Thus, these networks were not able to effectively capture minute variations and subtle changes in expressive regions. To address these issues, we introduce an active imaging concept to segregate active changes in expressive regions of a video into a single frame while preserving facial appearance information. Moreover, we propose a shallow CNN network: hybrid local receptive field based augmented learning network (OrigiNet) that efficiently learns significant features of the micro-expressions in a video. In this paper, we propose a new refined rectified linear unit (RReLU), which overcome the problem of vanishing gradient and dying ReLU. RReLU extends the range of derivatives as compared to existing activation functions. The RReLU not only injects a nonlinearity but also captures the true edges by imposing additive and multiplicative property. Furthermore, we present an augmented feature learning block to improve the learning capabilities of the network by embedding two parallel fully connected layers. The performance of proposed OrigiNet is evaluated by conducting leave one subject out experiments on four comprehensive ME datasets. The experimental results demonstrate that OrigiNet outperformed state-of-the-art techniques with less computational complexity.

*Keywords—OrigiNet, Hybrid feature block, Augmented Learning, RReLU, Active imaging, Micro expression recognition.*

## I. INTRODUCTION

Micro-expressions (MEs) represent the true emotions of a person and mainly arise when one is trying to surpass his/her actual intention behind manifested expressions (posed expressions) [1]. Although, MEs occur on persons' faces according to the emotions being experienced within a fraction of a second this emotional leakage exposes enough cues to understand the true feelings. Thus, MEs can be experienced only in a few frames as they last between 1/25 to 1/30 seconds [2][3]. Micro expression recognition (MER) is a challenging task due to low intensity and rapid movement of facial muscles. Firstly, Haggard et al. [4] discovered the micro-expressions in 1966. Later, Ekman et al. [5] also noticed micro-expressions, while investigating a video of a psychiatric patient. Furthermore, they developed a micro-expression training tool (METT) [6] to identify the micro expression dynamics. However, METT tool can achieve only 40% accuracy as observed by Frank et al. [7], which is insufficient for practical applications. ME analysis has attracted researchers as it has an indispensable role in various applications as police interrogation, depression analysis, forensic, entertainment, medical diagnosis, law enforcement etc. In literature many local binary pattern (LBP) [8] based descriptors as LBP in three orthogonal planes (LBP-TOP) [9], LBP with six intersection points (LBP-SIP) [10], LBP with mean orthogonal planes (LBP-MOP) [11], spatiotemporal completed local quantization patterns (STCLQP) [12], spatiotemporal LBP with integral projection (STLBP-IP) [13] and revisited integral projection (DiSTLBP-RIP) [14] were proposed to encode spatial and temporal changes from the micro expression video sequences. However, handcrafted feature descriptors fail to attain high accuracy due to the extraction of only superficial features and lack in eliciting sufficient knowledge for abstract features representation.

Recently, the adoption of deep networks such as AlexNet [15], VGG Net [16], ResNet [17], GoogleNet [18] and MobileNet [19] have created a tremendous take-off in the field of computer vision. The literature [20-22] in the field of MER, shows that convolutional neural networks (CNN) based networks also achieve impressive results up to some extent.

## II. LITERATURE STUDY

Conventionally, handcrafted feature extraction techniques were popular for MER. Wang et al. [23] introduced a tensor independent color space model (TICS) to spot the micro-expressions based on color components. They divided the video sequences into 4D structure: first 2D structure represents the spatial texture patterns, third dimension holds the momentary variations features and fourth dimension describes RGB color components. Moreover, Happy et al. [24] proposed a fuzzy histogram-based optical flow orientation technique (FHOFO) to capture temporal features of the micro-expressions. Zong et al. [25] introduced a hierarchical spatiotemporal descriptor, where discriminative features were extracted by dividing the input images into grids and classified using kernelized group sparse learning (KGSL). Moreover, Liu et al. [26] extended their work and proposed a sparse MDMO descriptor that utilized the classic graph regularized sparse coding scheme to extract prominent spatiotemporal features.

With the advancement in technology, deep learning approaches achieved high performance in the field of MER. Patel et al. [20] utilized the transfer learning approach over VGG-FaceNet and elicited the salient features from frames. Furthermore, an evolutionary search is applied to detect the disparities between the frames of micro expressions. Khor et al.

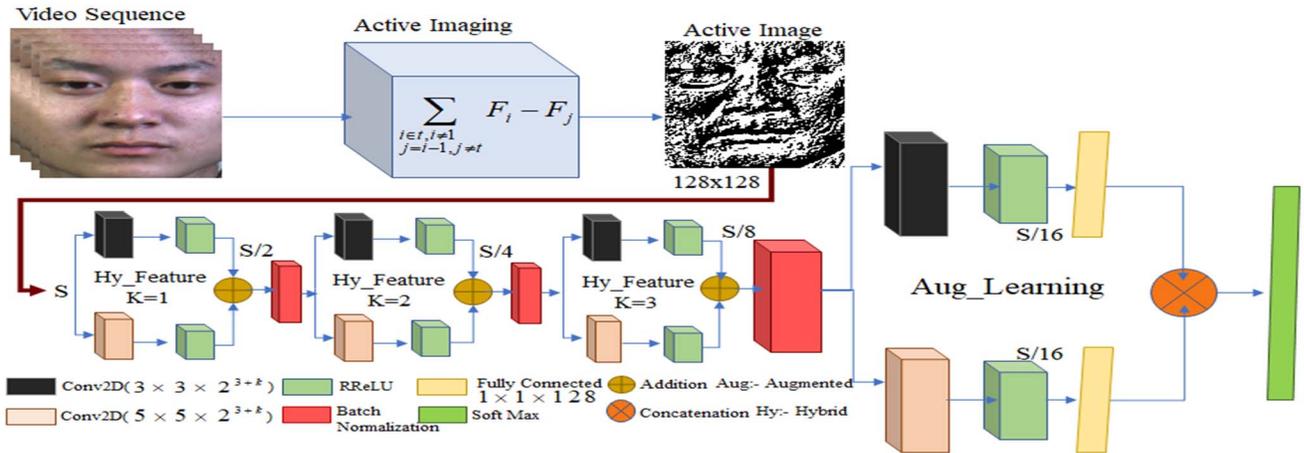

Fig. 1. The proposed framework for micro-expression classification.

[21] adopted CNN network and long short-term memory (LSTM) to learn the Spatio-temporal information for each image frame. Moreover, Li et al. [22] introduced a 3D flow CNN network, which incorporated optical flow information with CNN network to learn deep features of minute variation responsible to spot micro expression class. Liong et al. [27] also utilized optical flow to represent flow variations between frames and fed those to three parallelly connected CNN layer streams that learn the salient features of micro-expressions and classify them accordingly.

Based on the literature, CNN networks for MER systems can be divided into two categories: 2D-CNN with fixed-size image sequences and 2D-CNN with RNN or LSTM approaches. The first group of approaches used subsequences of videos with fixed size as input data. Thereby, these approaches allow 2D-CNN networks to work efficiently instead of 3D- CNN [28]. Although, these approaches are simple and successfully extract the local changes occurred within a small-time window, but they fail to capture long term motion features in long term video sequences. To solve these issues another type of network such as RNNs [29] and LSTMs [30] comes into literature. RNN networks hold memory cells to learn temporal features of the videos. Therefore, RNN networks are sensitive to both long- and short-term video sequences. Moreover, LSTMs were used with 2D-CNN networks. Where, 2D- CNN networks were utilized to elicit spatial features to represents the appearance structure of micro-expression video frames and LSTM block is used to encode the motion flow between frames. However, LSTMs based networks fail to establish a relationship between spatial and temporal features occurring simultaneously in frames [22]. In literature, some CNN networks also used predefined features like optical flow and saliency maps [26] to enriched the CNN model with flow information. However, the optical-flow, supply only the local dynamic features and simple summarization is performed to accumulate the local motion [28]. Furthermore, most of the existing CNN networks hold a sized filter throughout the network that allows capturing only a-like features only. Therefore, these networks unable to learn sufficient features and degrades the performance of the MER systems.

Our proposed work focuses on these problems. First, active imaging concept is introduced to handle the temporal features along with spatial features. Active images are computed to represent the Spatio-temporal information into one instance for each video. Furthermore, we proposed a 2D-CNN named OrigiNet, which elicits the dynamics of micro expression. The OrigiNet utilizes the strength of multi-scale convolution filters [18] to capture the local receptive field. Mainly OrigiNet adopts smaller size filters as active images hold minute variations to represent dynamic features of the video sequences. Furthermore, inspired by the study [31], we have introduced a new activation function named RReLU to improve the sparsity and convergence quality of the network by extending the range of derivatives as compared to ReLU variants. Furthermore, to enhance the learning capability of the network, we have used the augmented learning block. Augmented block merges responses of two different fully connected layers and reinforces the OrigiNet to learn two different segment features of the micro-expressions.

The performance of the proposed method is tested on three benchmark micro expression datasets: CASME-I [32], CASME-II [33], $CASME^2$ [34] and SAMM [35]. Experimental results and comparative analysis show that our method gains excellent performance as compared to the existing state-of-the-art methods.

III. PROPOSED METHOD

Most of the state-of-the-art approaches follow two-stage architectures as 2D CNN along with RNN [29] or LSTM [30]. However, these networks relied on complex functions and increasing number of parameters to capture the spatiotemporal features from the video sequence. Thus, we propose a candid and lightweight CNN based framework as shown in Fig. 1. First we compute single instances for each micro expression video using active imaging concept. Active imaging summarizes a video content into a single frame by preserving high stake active dynamics of the expressive regions. Further, these responses are fed to a CNN network: OrigiNet to extracts and learn the active expressive features from the active images.

A. Active Imaging

MEs are generated in high stake situations when a person trying to hide their true emotions and can be perceived only in a few frames. Thereby, MEs are rapid and subtle in nature thus hard to detect. In this paper, we introduced active imaging to represent the minute variations that arise momentarily. Active images encapsulate the spatial as well as temporal information of a video sequence into a single instance as shown in Fig. 2. From the figure, we can see that, motion changes between the frames are almost negligible, but active images can capture these changes as highlighted in yellow boxes. Yellow boxes highlighted that *in disgust: eyebrows, nose and lip corners, in repression: eyes, eyebrows and nose upper part, in surprise: eyes and in anger: glabella eyes, and mouth;* expressive regions are most affective.

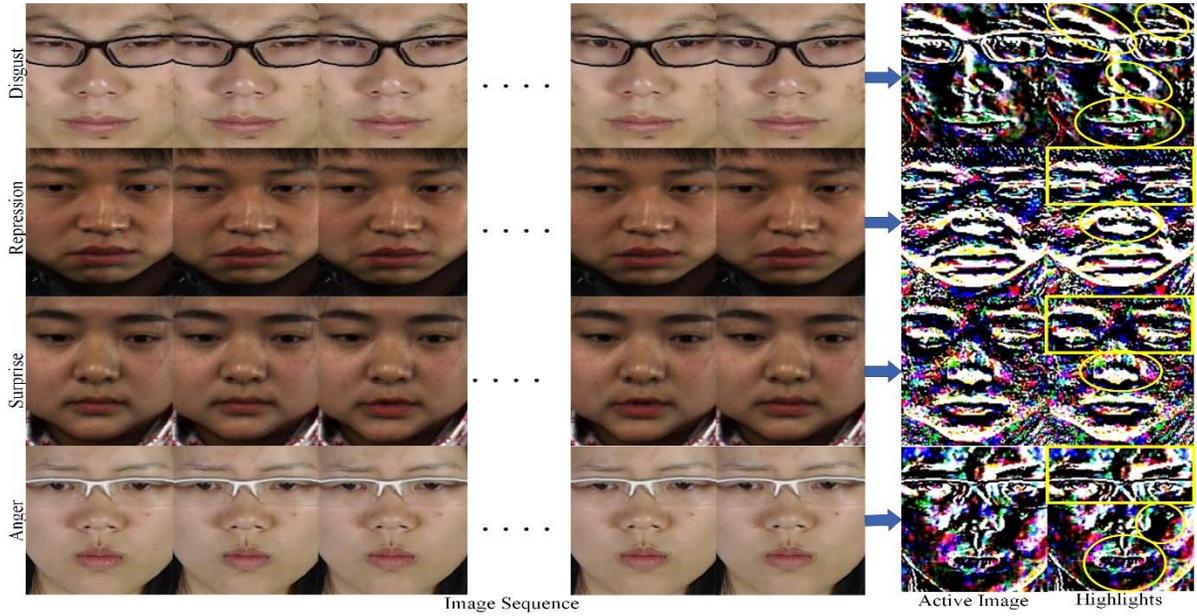

Fig. 2. Visual representation of active imaging for disgust, repression, surprise and anger micro expression video sequences.

The construction of active imaging is as follows. Let a video $V(F_\tau)$ holds $F_1, F_2, ... F_\tau$ frames. Here, we used time constraints ($\tau$) to assign a ranking $(1, 2, ..., \tau)$ of the frames. Since micro expressions are generated in a sequence (onset-apex-offset) thus later frames are assigned with large ranking. The active image is computed by Eq (1).

$$A = \sum_{t=2}^{\tau} E(t) \quad (1)$$

$$E(t) = FT(t-1) + FT(t) \quad (2)$$

$$FT(t) = \{F_i - F_j\}_{i \in t, i \neq 1, j = i-1, j \neq \tau} \quad (3)$$

In Eq. (1) $E(\bullet)$ is used to extract spatiotemporal features from image sequences. In first step, it extracts temporal movement features that occurred between frames w.r.t to each pixel component. Further, it cumulates the responses to preserve the appearance information as well. Finally, all features are aggregated to culminate the active image.

### B. OrigiNet

The proposed OrigiNet comprises three modules: 1) Hybrid feature block- to extract the features of facial appearance dynamics, 2) RReLU- to enhance the non-linearity by improving learning capability and 3) Augmented learning block- to boost the discriminative learnability of the network.

*Hybrid feature block*

Hybrid feature block contains laterally connected convolutional (Conv) layers with $3 \times 3$ and $5 \times 5$ sized filters along with activation function. These filters have a smaller receptive field to look at very few pixels at once. Thus, these filters allow the network to preserve micro-level edge variations that have a high impact in identifying the momentary changes in active images. Moreover, as per literature [36] deeper networks do not achieve high performance due to smaller datasets. Therefore, in the proposed network we have utilized three consecutive hybrid feature blocks and create a shallow network as micro expression datasets hold a comparatively smaller number of images for training.

*Refine Rectified Linear Unit (RReLU)*

In this paper, we have proposed an upgraded non-linear activation function named refined ReLU (RReLU) to improve the performance of the model. RReLU utilizes the functionalities of two activation functions: sigmoid and Leaky ReLU to incorporate non-linearity and sparsity in the network. Specifically, RReLU uses the capability of the ReLU to remove gradient saturation and alleviate the problem of vanishing gradient. Furthermore, to overcome the dying ReLU problem, RReLU exploits the robustness of sigmoid to handle negative values. Thus it, improves the optimization of CNN models during training. The RReLU refines the input feature maps and preserves only true edges. Particularly, sigmoid is applied to the input maps to decide the significance of features. Leaky ReLU is used to preserve intense edge features. Furthermore, resultant maps are multiplied to forget the trivial edge variations. Finally, response maps of sigmoid are integrated with previously updated maps to capture preserves effective variations. Let $A(x, y)$ be an input, then activation function for RReLU

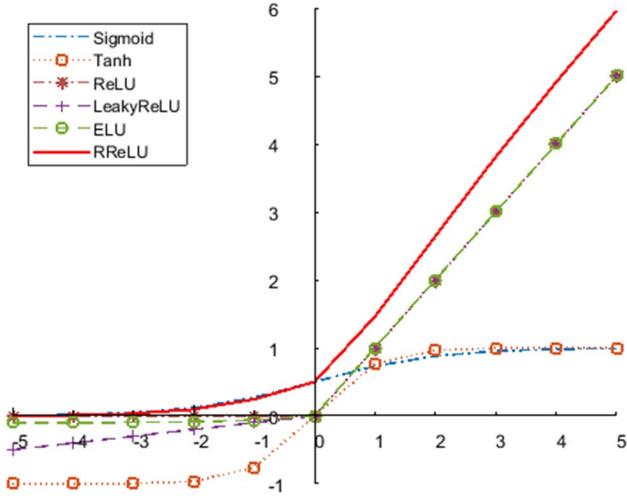

Fig. 3. Comparative study of the non-linear activations.

is computed by using Eq (4).

$$\psi(A(x,y)) = \begin{cases} \dfrac{e^{A(x,y)} + \alpha e^{A(x,y)} \cdot A(x,y)}{1 + e^{A(x,y)}} & A(x,y) < 0 \\ \dfrac{e^{A(x,y)} + e^{A(x,y)} \cdot A(x,y)}{1 + e^{A(x,y)}} & A(x,y) \geq 0 \end{cases} \quad (4)$$

Where $\alpha = 0.1$ is a constant value.

Conventionally sigmoid activation is employed in neural networks and shown impressive performance as compared to linear activation functions. However, sigmoid activation suffers from *vanishing gradient* problem because gradient values are saturated at 0 or 1 as shown in Fig. 3 by *blue dot dash* line. Thus, the network fails to learn significant features and degrades the performance of the CNN model. Similarly, tanh activation is proposed by scaling the sigmoid function between *[-1 1]* as shown in Fig. 3 highlighted with *orange dot o* line. tanh serves the responses' range nearly zeros, which leads to larger derivatives. Thus, tanh improves convergence of network and enhance the speed of the network. Moreover, tanh also suffers from vanishing gradient problem.

In recent time Nair et al. [37] proposed a rectifier linear unit (ReLU) activation that increase the sparsity by converting the values into *[0, x]* range, represented by *brown dash * line in Fig. 3. Moreover, ReLU also resolves the problem of vanishing gradient and extends the robustness of the model. However, ReLU is saturated with zeros for negative values that generate the problem of *dying ReLU* and deteriorate the optimization performance of the network. Further, Maas et al. [38] and Clervert et al. [39] introduced an extended version of ReLU as Leaky ReLU and exponential linear unit (ELU) respectively. Leaky ReLU and ELU solve the problem of dying ReLU by activating negative values as shown in Fig. 3, highlighted by *purple dash +* and *green dash o* line. However, both Leaky ReLU and ELU have a slower convergence rate due to a small range $[\alpha,1]$ and $[\alpha e^x,1]$ of derivatives, respectively. Therefore, to overcome this issue and improve the learnability of the model we propose RReLU. The RReLU is a restricted network to vanishing gradient and dying ReLU problem as demonstrated in Fig. 3 by *dark red line* and has wide range $[\alpha \times x, x]$ derivatives. Thus, RReLU increases convergence and enhances the learnability of the model. Detail sheet of non-linear functions is tabulated in Table I.

*Augmented learning block*

Motivated by the literature [40] [41], in this network we introduced an augmented learning block that allows network to learn more discriminative semantics (position, orientation etc.) invariant features. Augmented learning block holds two parallelly connected FC layers to yield global configuration from different scaled features that are preserved by previous laterally connected layers. Moreover, resultant features are jointly forwarded to next classification layers to learn discriminative features of micro expressions.

Let $I(l,m)$ be an input image and $\mathbb{C}_S^{x,y,N}(\bullet)$ represents Conv function, where, S implies for stride, N is depth, x and y stand for the size of the filter. Then, output of network is computed by Eq (5-11).

TABLE I. ACTIVATION FUNCTIONS DETAILS SHEET.

| Activation | Forward Pass | | Backward Pass | |
|---|---|---|---|---|
| | *Equation* | *Range* | *Derivation* | *Range* |
| **Sigmoid** | $S(x) = \sigma(x)$ | $[0,1]$ | $S'(x) = S(x)(1-S(x))$ | $[0,1]$ |
| **Tanh** | $T(x) = (2 \times \sigma(2x)) - 1$ | $[-1,1]$ | $T'(x) = (1 - T(x)^2)$ | $[0,1]$ |
| **ReLU** | $R(x) = \begin{cases} 0 & x < 0 \\ x & x \geq 0 \end{cases}$ | $[0,x]$ | $R'(x) = \begin{cases} 0 & x < 0 \\ 1 & x \geq 0 \end{cases}$ | $[0,1]$ |
| **ELU** | $E(x) = \begin{cases} \alpha(e^x - 1) & x < 0 \\ x & x \geq 0 \end{cases}$ | $[-\alpha(1-e^x), x]$ | $E'(x) = \begin{cases} E(x) + \alpha & x < 0 \\ 1 & x \geq 0 \end{cases}$ | $[\alpha e^x, 1]$ |
| **Leaky ReLU** | $LR(x) = \begin{cases} \alpha x & x < 0 \\ x & x \geq 0 \end{cases}$ | $[\alpha x, x]$ | $LR'(x) = \begin{cases} \alpha & x < 0 \\ 1 & x \geq 0 \end{cases}$ | $[\alpha, 1]$ |
| **RReLU** | $RR = \begin{cases} \dfrac{e^x + \alpha e^x \cdot x}{1 + e^x} & x < 0 \\ \dfrac{e^x + e^x \cdot x}{1 + e^x} & x \geq 0 \end{cases}$ | $[0, x+1]$ | $RR'(x) = \begin{cases} \dfrac{e^x(e^x + \alpha x + 2)}{(e^x + 1)^2} & x < 0 \\ \dfrac{e^x(e^x + x + 2)}{(e^x + 1)^2} & x \geq 0 \end{cases}$ | $[\alpha x, x]$ |

$$E_{mt} = S_{fm}\left(FC^{128}(\xi) \| FC^{128}(\xi)\right) \quad (5)$$

Where, FC represents the fully connected layer.

$$\xi = BN\left(\Re\left(\mathbb{C}_2^{3,3,96}(\delta(3))\right) + \Re\left(\mathbb{C}_2^{5,5,96}(\delta(3))\right)\right) \quad (6)$$

$$\delta(k) = \begin{cases} I(l,m) & k=0 \\ BN\left(\Re\left(\mathbb{C}_2^{3,3,d(k)}(\delta(k-1))\right) + \Re\left(\mathbb{C}_2^{5,5,d(k)}(\delta(k-1))\right)\right) & k \geq 0 \end{cases} \quad (7)$$

Where,

$$d(k) = 2^{3+k} \quad (8)$$

$$\mathbb{C}_S^{x,y,N}(I(l,m)) = \sum_{i=-y/2}^{y/2} \sum_{j=-x/2}^{x/2} f_k(i,j) \otimes I(\alpha-i, \beta-j) \quad (9)$$

$$\begin{cases} \alpha = (S \times l - (S-1)) \\ \beta = (S \times m - (S-1)) \end{cases} \quad (10)$$

Here, $f_k$ represents the convolutional kernel.

$$S_{fm}(p_s) = \frac{e^{p_s}}{\sum_{t=1}^{Z} e^{p_t}} \quad (11)$$

Where $t \in Z$ represents the dimension of the resultant feature vector.

## C. Comparative study

It is hard to extract prominent features from the video sequences as compared to a single image. Thus, existing MER approaches [21-22] follow a complex process to capture spatiotemporal features of micro-expressions from the video sequences. Moreover, most of the benchmark datasets contain unequal frame numbers since existing methods [23-25] use interpolation to normalize them. However, interpolation takes some time to compress the data and loses essential information. In this paper, we have proposed an active imaging concept that represents the whole video information into a single image without any interpolation. Thus, it overcomes the complications of handling video sequences. Moreover, we have proposed a portable shallow OrigiNet to capture actively changing features. Although, in literature, various CNN networks like VGG Net [16], ResNet [17] and Mobile Net [18] are proposed to learn features and classify the objects accordingly. But these CNN networks are not suitable to capture micro-level edge variations. They utilize sequentially coupled Conv layers with identical filter sizes that mainly focus to capture homogenous scaled receptive field and avoid fine-tuned edge variations. Thus, these networks lack in gathering adequate learning of spatial structure that degrades the performance of MER system. Moreover, some of the existing networks like Google Net [18] and NIN [42] exploit the functionality of multi-scaled Conv filters that allows a network to capture extensive as well as small space features. Though, these networks achieve impressive results but increase complexity and require a large pool of memory, which is not suitable for smart and handheld devices. Furthermore, we have observed that smaller filter sizes are more preferable for micro-expression recognition. Kernel with large scales (7× 7 and 11× 11) have a larger receptive field per layer and allow to extract generic features spread across the image. Therefore, these filters focus on abstract transitional information and skip the minute information, which is quite important in micro expression. The learning capability of OrigiNet is demonstrated in Fig. 4. From the fig. we can validate that OrigiNet extract minute (highlighted with red boxes) but influential edge variations of the micro-expressions.

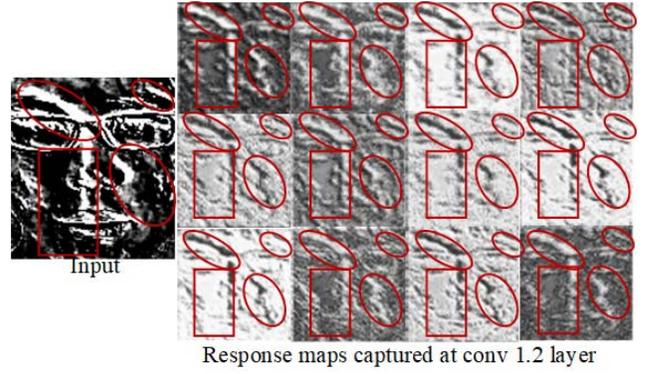

Fig. 4. Visualization of neurons responses captured at Conv 1$^{st}$ layer. (Visualized responses are mean response of all collected neurons).

Furthermore, in this paper we have used proposed RReLU activation instead of regular ReLU function to avoid the gradient saturation and zero gradient issues in optimization. RReLU enhances sparsity of gradient and lies in a range [0, x+1]. Moreover, RReLU also increases convergence speed as its derivative range is greater. Furthermore, OrigiNet does not include a pooling layer for downsampling unlike the other CNN based models. Some studies in literature [43-45] have shown, that replacing pooling by convolution with more strides adds inter-feature dependencies. It also learns the pooling operation by parameter sharing among channels rather than fixing it. Thereby, we have incorporated convolution with stride 2 that outperforms the pooling operation parameter sharing among channels rather than fixing it. Thereby, we have incorporated convolution with stride 2 that outperforms the pooling operations and improves the network learning ability for the micro-expression recognition.

## IV. EXPERIMENTAL SETUP

To validate the effectiveness of proposed OrigiNet, we have performed four experiments over on benchmark datasets: CASME-I [32], CASME-II [33] CSAME$^2$ [34] and SAMM [35]. These datasets are prepared to analyze the spontaneous micro-expressions under challenging scenarios like ethnicity variations, subjects wearing different artifacts, gender variations etc. In the literature [23], mainly two types of validation schemes are used to test the efficiency of MER systems: leave one video out (LOVO) and leave one subject out (LOSO) cross-validation. Study [16] found that LOVO cross-validation executed the results in person dependent manner that is not much feasible to validate MER system as it has a dependency on subjects. Thus, in this paper, all experiments are computed by using LOSO strategy, where one subject is used for testing and remaining are considered as a training image set. Furthermore, to arrange the training and validation sets, training dataset is randomly divided into two parts with a ratio of 80:20 respectively. Moreover, SAMM dataset has row images since we have used the Viola Jones face detection algorithm [46] to extract the region of interest. In our experiments, we augmented the generated active images and create a large pool of data to avoid the problem of

TABLE II. THE NO. ACTIVE IMAGES USED FOR DIFFERENT DATASETS IN THE EXPERIMENTAL RESULTS.

| Emotion | CASME-I | CASME-II | CASME$^2$ | SAMM |
|---|---|---|---|---|
| Anger | - | - | 102 | 57 |
| Contempt | 1 | - | - | 12 |
| Disgust | 43 | 58 | 88 | 9 |
| Fear | 2 | 2 | - | 8 |
| Happy | 12 | 32 | 151 | 27 |
| Others | - | 99 | - | 25 |
| Sad | 5 | 7 | - | 6 |
| Surprise | 21 | 25 | - | 15 |
| Repression | 38 | 27 | - | - |
| Tension | 67 | - | - | - |
| Total | 189 | 250 | 341 | 159 |

TABLE III. RECOGNITION ACCURACY COMPARISON OF CASME-I AND CASME-II DATASET.

| Method | Task | CASME-I | CASME-II |
|---|---|---|---|
| STLBP-IP* [12] | (H, S, D, R, O) | N/A | 59.91 |
| TICS* [23] | (P, N, S, O) | 61.86 | 61.11 |
| FHOFO* [24] | (P, N, S, O) | 65.99 | 55.86 |
| CNN-LSTM* [30] | (H, S, D, R, O) | 60.98 | N/A |
| 3D-Flow* [22] | (H, S, D, R, T) | 55.44 | 59.11 |
| VGG-19 [16] | (P, N, S, O) | 62.26 | 53.30 |
| MobileNet [19] | (P, N, S, O) | 65.68 | 57.72 |
| MobileNet V2 [19] | (P, N, S, O) | - | 60.76 |
| **OrigiNet** | (P, N, S, O) | **66.30** | **62.09** |

* This result is from the corresponding original paper and H, S, D, R, T, P, N, O stands for Happy, Surprise, Disgust, Repression, Tense, Positive, Negative, Others.

overfitting in training. Specifically, each active image is rotated between [-45°,45°] degree with an increment of 15° and enhanced by performing histogram equalization.

*A. Experiments on CASME-I dataset*

The Chinese Academy of Sciences Micro-expression (CASME) [32] dataset contains 1500 facial movements and 195 spontaneous micro-expressions recorded by 60 fps camera with 1280×720 resolution. CASME-I dataset consists of 19 subjects' micro-expressions annotated with eight emotion classes named contempt, tense, disgust, happiness, surprise, fear, sadness and repression. However, in CASME-I dataset, some emotions like fear, sadness and contempt hold very few samples and some of emotion labels are ambiguous. Thus, most of the existing approaches [12-14] dropped these emotion classes to balance the inequality issue in datasets. Recently, some methods [23-25] created new emotion classes by merging the existing emotions as positive: *happy*, negative: *disgust, sad, fear*, surprise and other: *contempt, repression, tense*. In our experimental setup we have utilized the merged emotion classes and finally gather 189 active images as: positive: 12, negative: 50, surprise:21 and others: 106. A detailed description of dataset samples is tabulated in Table II. Table III illustrates the effectiveness of the proposed OrigiNet and other existing methods in terms of recognition accuracy. From Table III, it is clear that the proposed model outperforms the other existing models with comfortable margins. More specifically, it secures 4.04% and 0.62% improvement as compared to the CNN based networks: VGG-19 and MobileNet respectively. From results, it is clear that former models are not achieving high accuracy due to deep networks and large channel depths. OrigiNet also obtains 4.44% and 0.31% more recognition rate as compared to handcrafted descriptors of TICS and FHOFO respectively.

TABLE IV. RECOGNITION ACCURACY COMPARISON OF CASME$^2$ AND SAMM DATASET.

| Method | CASME$^2$ | SAMM |
|---|---|---|
| LBP-TOP-SVM* [34] | 40.95 | N/A |
| VGG-19 [16] | 54.10 | 30.09 |
| MobileNet [19] | 53.20 | 31.84 |
| **OrigiNet** | **54.56** | **34.89** |

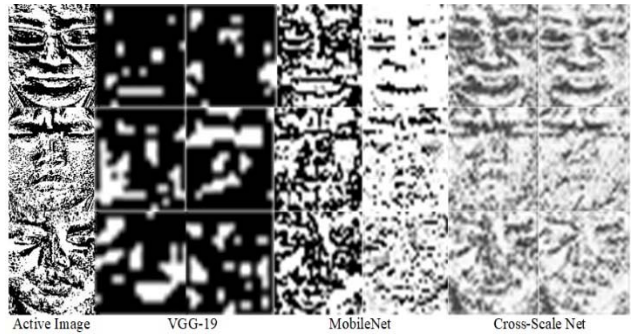

Fig. 5. Visual representation state-of-the-art models: VGG-19, MobileNet and OrigiNet over three different emotions a) Negative, b) Other and Positive. (Visualized responses are mean response of all collected neurons)

*B. Experiments on CASME-II dataset*

The CASME II [33] is an improved spontaneous micro-expressions database which has a higher temporal resolution (200 fps) and spatial resolution (about 280×340 pixels on facial area). It consists of 247 spontaneous micro-expressions of 26 subjects. The dataset is prepared in a well-arranged environment with proper lighting to avoid the problem of illumination flickering. CASME-II dataset holds seven expression categories as disgust, fear, happiness, other, repression, sadness, surprise. Similar to CASME-I, we also convert CASME-II dataset into four categories: positive: *happy*, negative: *disgust, fear, sad*, others: others, *repression* and surprise. Finally, we have collected 250 active images as positive: 32, negative: 67, surprise: 25 and others: 126. Table III, shows the effectiveness of the proposed OrigiNet over existing state-of-art methods in terms of recognition accuracy. From Table III, it is clear that the proposed network obtains higher recognition accuracy as compared to other approaches. It achieves performance gain of 8.79%, 4.37% and 1.33% over VGG-19, MobileNet and MobileNet V2 respectively. It also achieves 0.98% and 6.23% more accuracy in comparison to traditional handcrafted techniques TICS and FHOFO.

*C. Experiments on CASM$^2$ dataset*

The CASME$^2$ [34] contains 300 spontaneous macro expression samples and 53 micro-expression samples recorded at 30 fps camera. It consists of 22 subjects and 3 basic categories of micro-expressions (i.e. happy, anger, disgust). In our experimental setup, we have selected a total 341 image sequences: anger-101, happy-151 and disgust-89 of micro expressions. Table IV demonstrates the improved accuracy results obtained by proposed OrigiNet as compared to existing approaches. Specifically, the proposed method obtained 13.61%, 0.46% and 1.36% more accuracy as compared to base results (LBP-TOP), VGG-19 and MobileNet respectively. Results in Table IV validate the effectiveness of the proposed network as compared to the other existing state-of-the-art approaches.

TABLE V.  COMPUTATIONAL COMPLEXITY ANALYSIS OF ORIGINET AND EXISTING NETWORKS.

| Network | # Layers | # Parameters (in millions) | #Memory (in megabytes) |
|---|---|---|---|
| AlexNet [15] | 8 | 61 | 227 |
| VGG-16 [16] | 16 | 138 | 515 |
| VGG-19 [16] | 19 | 144 | 535 |
| GoogleNet [17] | 22 | 7 | 27 |
| ResNet 50 [18] | 50 | 25 | 98 |
| MobileNet [19] | 20 | 4 | 16 |
| MobileNet V2 [19] | 53 | 3.5 | 14 |
| **OrigiNet** | **10** | **1.8** | **14.3** |

## D. Experiments on SAMM dataset

The Spontaneous Actions and Micro-Movements (SAMM) [35] is a newer database recorded at 200 fps and resolution set to 2,040×1,088 pixels. It consists of 29 subjects and with eight identified categories of expressions (i.e. Other, Disgust, Happiness, Contempt, Fear, Sadness, Surprise, Anger). Table IV represents the effectiveness of proposed network over existing approaches. Particularly, the proposed method obtained 4.8%, 3.05% more accuracy as compared to VGG-19 and MobileNet respectively. Results in Table IV prove that the OrigiNet outperforms the other existing state-of-the-art approaches.

## E. Computational complexity analysis

This section provides a comparative analysis of the computational complexity between the existing and proposed network. The total number of parameters involved in each network are tabulated in Table V. The proposed OrigiNet has only 1.8 million learnable parameters which are very less as compared to other existing benchmark models like: AlexNet:61M, VGG-16: 138M, VGG-19: 144M, GoogleNet: 7M, ResNet: 25M mobileNet: 4M and MobileNet V2: 3.5M. Moreover, OrigiNet architecture has fewer depth channels and hidden layers as compared to former methods. Particularly, OrigiNet comprises of 10 layers. In comparison to that, VGG-16, VGG-19, GoogleNet and ResNet consist of 16, 19, 22 and 34 layers respectively. Furthermore, OrigiNet taking only 14.3 MB memory storage that is very less as compare to GoogleNet: 27, ResNet: 44 MB and Mobile Net 16 MB.

## F. Qualitative analysis

The learning capability of proposed network as compared to state-of-the-art is shown in Fig. 5. Fig. 5. demonstrates the two most effective visual representations of different emotion classes as *negative, other and positive* from all CASME-2 dataset. From figure, it is clear that the response feature maps significantly assist in preserving the dynamic variations in different expressive regions in the facial image. For example, *in negative: eyes, eyebrows, mouth* regions; *in other: glabella and in positive: eyes, mouth;* give maximum affective response for related facial expressions. Therefore, we can conclude that proposed network has preserved more relevant feature responses and outperform the existing CNN based networks VGG-19 and MobileNet for almost all emotion classes.

## G. Ablation Study

To analyses the effect of each block of OrigiNet such as active imaging, RReLU activation, hybrid feature block and augmented learning block, we conducted six ablation experiments over CASME-II datasets. First, to analyses the role of active imaging, we have evaluated OrigiNet over Dynamic Imaging [28] named OrigiNet-Dy and results are tabulated in Table- VI. Second, to examine the effect of RReLU over Sigmoid and Leaky ReLU activation functions, we have conducted two experiments named OrigiNet sig and OrigiNet LR respectively. Further, we compute results by replacing parallel-connected FC layers by linearly connected single FC layer named as OrigiNet. Results tabulated in Table-VI proved the empirical effect of parallel-connected FC layers in augmented learning block. Moreover, to test the effect of kernel sizes in Hybrid feature block, we evaluated the results with kernel sizes ($1 \times 1$ and $3 \times 3$) and ($5 \times 5$ and $7 \times 7$) in place of ($3 \times 3$ and $5 \times 5$) named as OrigiNet KS-1 and OrigiNet KS-2, respectively. Table VI validated the significant importance of each module of OrigiNet.

TABLE VI.  ABLATION STUDY ANALYSIS OVER CASME-II DATASET IN TERMS OF RECOGNITION AND COMPUTATIONAL COMPLEXITY.

| Method | Task | Acc. | # Param. | #Mem. |
|---|---|---|---|---|
| OrigiNet Sig | (P, N, S, O) | 54.13 | 1.8M | 14.3MB |
| OrigiNet LR | (P, N, S, O) | 57.79 | 1.8M | 14.3MB |
| OrigiNet LFC | (P, N, S, O) | 52.92 | 0.9M | 7.6MB |
| OrigiNet KS-1 | (P, N, S, O) | 58.65 | 1.6M | 13.4MB |
| OrigiNet KS-2 | (P, N, S, O) | 59.61 | 2.2M | 17.9MB |
| OrigiNet WDy | (P, N, S, O) | 46.19 | 1.8M | 14.3MB |
| **OrigiNet** | (P, N, S, O) | **62.09** | 1.8M | 14.3MB |

*Here, Param., Mem., M and MB represent parameters, memory, millions and megabyte, respectively.*

## V. CONCLUSION

In this paper, we propose an active imaging-based OrigiNet that extracts the discriminative feature to classify MEs. First, active images are generated from the video sequence that conserves temporal information with appearance into a single frame. Furthermore, active images are forwarded to the OrigiNet for training and inference. OrigiNet contains two-scale filters such as $3 \times 3$ and $5 \times 5$ to extract pertinent features of the receptive field. Moreover, we have used RReLU that increases the divergence scale and enhances the learnability of hidden neurons. Furthermore, resultant feature maps are fed to the augmented learning block that holds laterally connected FC layers and combines the different scaled features. Thus, laterally connected FC layers allow elicitation of two different configurations and improve the discriminability of the network. Apart from that, proposed network has a reduced number of parameters so that it can be easily deployable in handheld and mobile gadgets. Moreover, experimental results validate the effectiveness of proposed network over state-of-the-art approaches.